%% file: main.tex
\documentclass[runningheads]{llncs}

\usepackage{accv}

\usepackage{accvabbrv}

\usepackage{graphicx}
\usepackage{booktabs}

\usepackage[accsupp]{axessibility}  

\usepackage{ascii}
\usepackage[utf8]{inputenc} 
\usepackage[T1]{fontenc}    
\usepackage{url}            
\usepackage{booktabs}       
\usepackage{amsfonts}       
\usepackage{nicefrac}       
\usepackage{microtype}      
\usepackage{float}

\usepackage{enumerate}
\usepackage{ntheorem}
\usepackage{makecell}

\usepackage{colortbl}
\usepackage{enumitem}
\usepackage{wrapfig}
\usepackage{amsmath}
\usepackage{amssymb}
\usepackage{mathrsfs}
\usepackage{multirow}
\usepackage{textcomp}
\usepackage{footnote}
\usepackage{soul}
\usepackage{threeparttable}
\usepackage[ruled,vlined]{algorithm2e}
\usepackage{color}
\definecolor{mygray}{gray}{.9}
\definecolor{light-gray}{gray}{0.5}
\definecolor{pretty-blue}{RGB}{0, 113, 188}
\definecolor{linecolor1}{gray}{.95} 
\definecolor{linecolor}{gray}{.895} 
\definecolor{kaiming-green}{RGB}{57,181,74} 
\definecolor{icmlblue}{rgb}{0,0.08,0.45} 
\def\eg{{\it{e.g.}}}

\def\ie{{\it{i.e.}}}

\usepackage{xfrac}
\usepackage{tabularx}
\usepackage{tikz}
\usepackage{bbm}

\usepackage[pagebackref,
    breaklinks,
    colorlinks,
    citecolor=accvblue
    ]{hyperref}
\usepackage{orcidlink}
\begin{document}

\title{Bringing Masked Autoencoders Explicit Contrastive Properties for Point Cloud Self-Supervised Learning}

\titlerunning{Point-CMAE}
\author{
Bin Ren$^{1,2}$\quad
Guofeng Mei$^{3}$\quad
Danda Pani Paudel$^{4,5}$\quad
Weijie Wang$^{2,3}$\quad
Yawei Li$^{5}$\\
Mengyuan Liu$^{6}$\quad
Rita Cucchiara$^{7}$\quad
Luc Van Gool$^{4,5}$\quad
Nicu Sebe$^{2}$ \\
$^1$University of Pisa,
$^2$University of Trento,
$^3$FBK,
$^4$INSAIT,
$^5$ETH Z\"urich,
$^6$Peking University,
$^7$University of Modena and Reggio Emilia\\
}
\institute{}




\maketitle

\input{sections/0_abstract}
\input{sections/1_introduction}

\input{sections/2_relatedworks}
\input{sections/3_methods}
\input{sections/4_experiments}
\input{sections/5_conclusion}

\appendix
\input{sections/appendix}

%
%
\clearpage
\bibliographystyle{splncs04}
\bibliography{main}
\end{document}

%% file: sections/0_abstract.tex
\begin{abstract}
    Contrastive learning (CL) for Vision Transformers (ViTs) in image domains has achieved performance comparable to CL for traditional convolutional backbones. However, in 3D point cloud pretraining with ViTs, masked autoencoder (MAE) modeling remains dominant. This raises the question: Can we take the best of both worlds?
    To answer this question, we first empirically validate that integrating MAE-based point cloud pre-training with the standard contrastive learning paradigm, even with meticulous design, can lead to a decrease in performance.
    To address this limitation, we reintroduce CL into the MAE-based point cloud pre-training paradigm by leveraging the inherent contrastive properties of MAE. Specifically, rather than relying on extensive data augmentation as commonly used in the image domain, we randomly mask the input tokens twice to generate contrastive input pairs. 
    Subsequently, a weight-sharing encoder and two identically structured decoders are utilized to perform masked token reconstruction. 
    Additionally, we propose that for an input token masked by both masks simultaneously, the reconstructed features should be as similar as possible. 
    This naturally establishes an explicit contrastive constraint within the generative MAE-based pre-training paradigm, resulting in our proposed method, Point-CMAE. Consequently, Point-CMAE effectively enhances the representation quality and transfer performance compared to its MAE counterpart. 
    Experimental evaluations across various downstream applications, including classification, part segmentation, and few-shot learning, demonstrate the efficacy of our framework in surpassing state-of-the-art techniques under standard ViTs and single-modal settings. The source code and trained models are available at~\url{https://github.com/Amazingren/Point-CMAE}.
  \keywords{Point Cloud Pre-training \and Self-Supervised Learning \and Vision Transformer}
\end{abstract}

%% file: sections/1_introduction.tex
\section{Introduction}
\label{sec:introduction}
\begin{figure}[!t]
    \centering
    \includegraphics[width=1.0\linewidth]{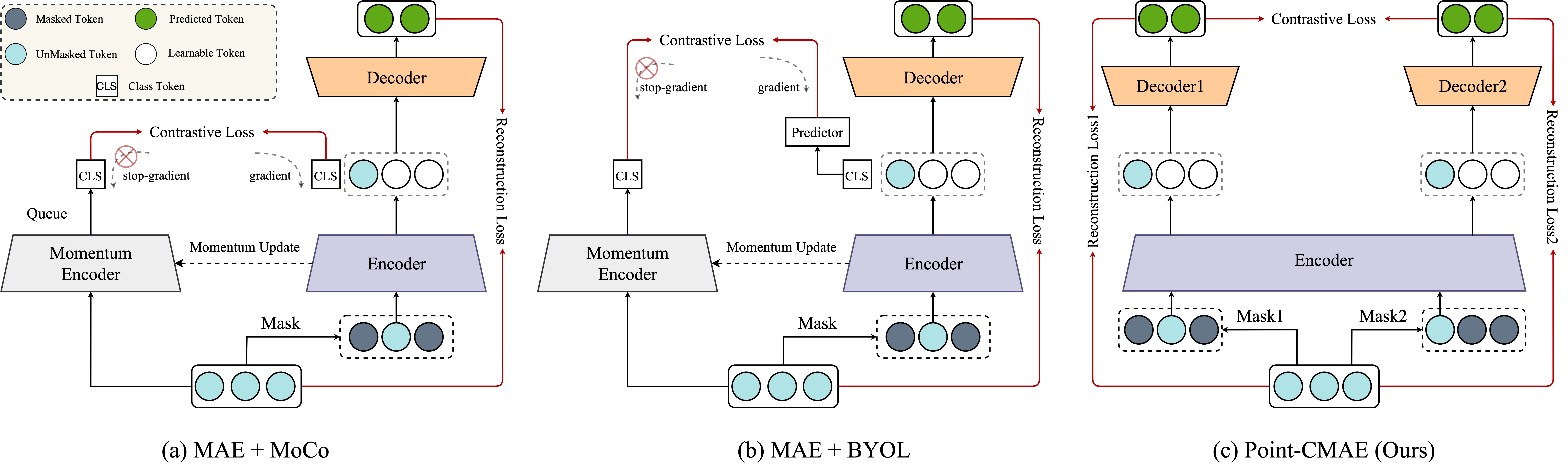}
    \caption{Illustration of: (a) MAE with MoCo~\cite{he2020momentum} style contrastive learning. An extra queue is required to maintain the negative samples during the entire pre-training. (b) MAE with BYOL~\cite{grill2020bootstrap} style contrastive learning. The asymmetric structure design with the fully connected predictor layer is used to exclude the negative samples. (c) The proposed Point-CMAE framework. 
    Two identically structured decoders, updated differently, are employed to introduce explicit contrastive properties within the generative self-supervised pre-training paradigm (\ie, MAE-based).}
    \vspace{-5mm}
    \label{fig:compare_demo}
\end{figure}
Understanding 3D scenes is critical for diverse applications, from autonomous vehicles navigating urban environments to robotic manipulation tasks~\cite{an2024rethinking,cai2023spsd}. Point clouds, representing objects and surroundings precisely, offer an advantage due to their easy acquisition and accurate geometry capture~\cite{mei2022overlap}. 
However, annotating 3D point cloud data is even more resource-intensive compared to image data, as each point requires labeling.
Self-supervised learning (SSL) has emerged as a key solution, the effectiveness of which has already been validated in natural language processing~\cite{devlin2018bert,radford2018improving,brown2020language}, computer vision~\cite{he2020momentum,chen2020simple,chen2021exploring,zhao2024denoising}, and multimodal learning~\cite{radford2021learning,zhang2022pointclip,jia2021scaling}, enabling informative feature representations from unlabeled input data. 
As a result, it is widely utilized in pre-training for 3D point clouds, demonstrating comparative or even better performance compared to its supervised counterpart in downstream tasks such as classification, segmentation, and detection~\cite{PointBERT, MaskPoint,qi2023contrast,ren2023masked,liu2020grouped}.

Specifically, contrastive (single/cross-modal) and generative-based (reconstruct/predict) pre-training strategies are widely employed. CL is renowned for capturing global information to improve the model's discriminative ability, and this has been demonstrated to be effective in various domains such as 2D images and point cloud pre-training methods based on convolutional neural networks (CNNs)~\cite{mei2024unsupervised}. 
However, deploying single-modal CL to point clouds with ViT often results in inferior performance compared to MAE-based pre-training. On the other hand, using MAE with a Chamfer constraint for point cloud reconstruction can lead to sub-optimal solutions~\cite{MaskPoint}, as ground truth points are sampled from the original point clouds, causing inconsistencies when forcing predicted points to match them.
In addition, the primary concept behind MAE typically emphasizes local relationships. 
These challenges collectively hinder the overall performance of ViTs-based point cloud pre-training. Consequently, a natural question arises: \textit{Can we leverage the strengths of both paradigms}? In other words, can we harness the power of contrastive learning to learn consistent and abstract feature representations, while also utilizing MAE to enhance the model's ability to capture local data structures? 
To explore this question, we propose to enrich the MAE paradigm with contrastive learning ability, where the masked and the unmasked point cloud tokens are treated as two augmentations. 
\begin{wrapfigure}[14]{r}{7cm}
    \centering
    \vspace{-4mm}
    \includegraphics[width=1.0\linewidth]{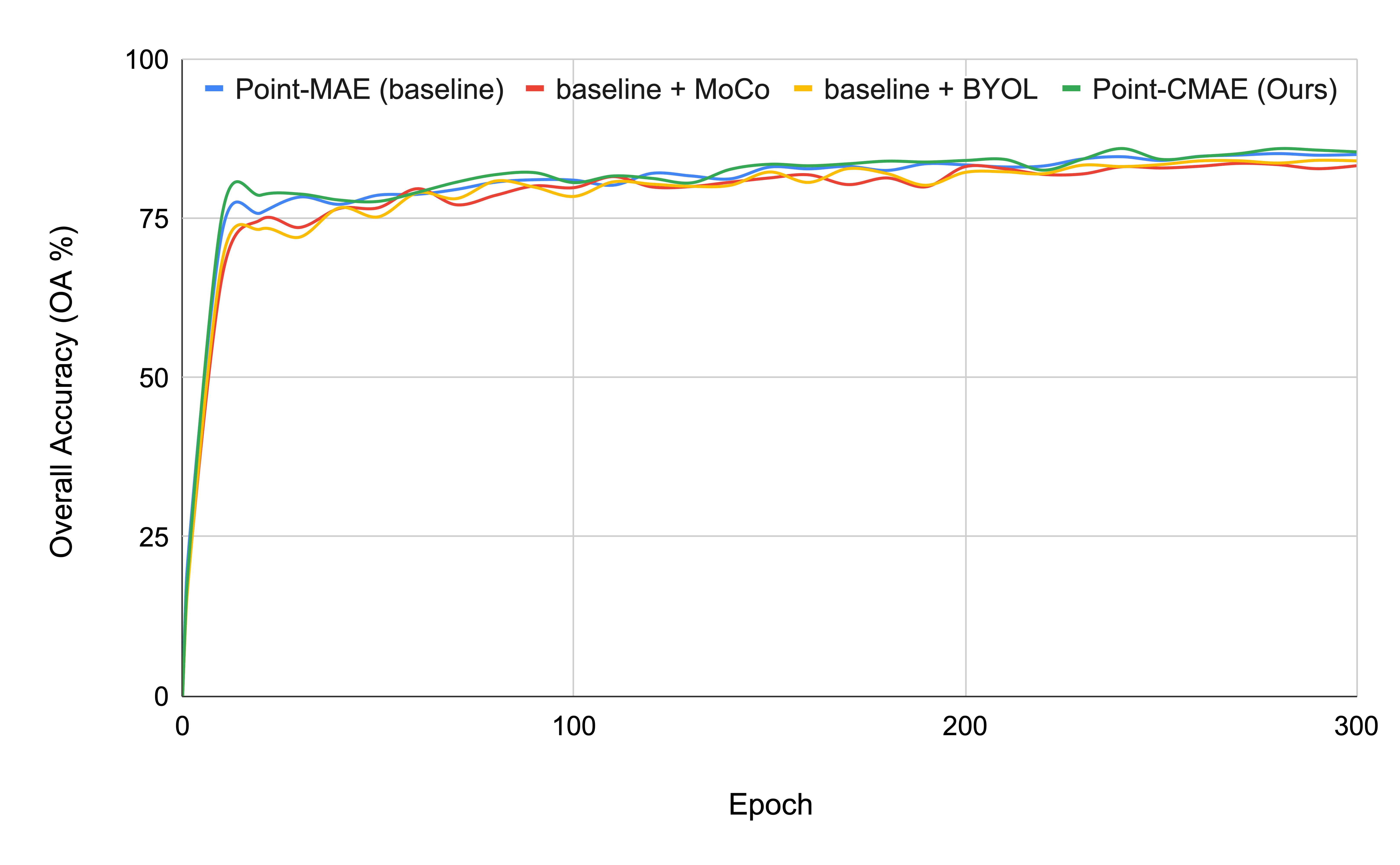}
    \vspace{-6mm}
    \caption{
    The classification comparison of different contrastive learning pipelines, integrated with the baseline method Point-MAE~\cite{PointMAE}, is conducted on the ScanObjectNN~\cite{uy2019revisiting} datasets.
    }
    \label{fig:contras_fig}
\end{wrapfigure}
Based on this setting, we deploy MoCo~\cite{he2020momentum}, and BYOL~\cite{grill2020bootstrap} style CL with Point-MAE~\cite{PointMAE} to point cloud pre-training with ViTs structure. 
The illustration of these two kinds of MAE-based CL is shown in Fig.~\ref{fig:compare_demo} (a) and Fig.~\ref{fig:compare_demo} (b), respectively. 
Our experimental results, as shown in Fig.~\ref{fig:contras_fig}, confirm that this approach frequently results in the occurrence of a performance drop.
This is primarily because making CL effective for ViTs-based point cloud pre-training necessitates a well-designed framework and a tailored approach to conducting separate data augmentation for point cloud data. These requirements significantly increase the complexity of the entire problem, highlighting the non-trivial nature of even bringing contrastive learning to an MAE-based paradigm.

To overcome the above limitations, we propose Point-CMAE, a simple yet effective method that explicitly integrates contrastive properties into the MAE-based point cloud pre-training paradigm using a ViTs architecture. The main idea of the proposed Point-CMAE is simplified in Fig.~\ref{fig:compare_demo} (c). Specifically, for a given point cloud token, we mask it twice randomly to construct the contrastive input pairs instead of applying heavy data augmentation. We then perform MAE for each of the masked input tokens using a weight-sharing encoder and two identically structured decoders. In this design, each decoder independently recovers the masked point cloud input token, which naturally forces the encoder to learn more common and representative features that satisfy the requirements of both decoders.
Though this largely improves the semantic understanding of the encoder, the reconstruction constraint, the Chamfer distance loss, is usually used to minimize the distance between the predicted masked points and the ground truth points regarding the fact the ground truth points are also just one sampling of the original point cloud, which naturally leads to a sub-optimal problem. 
Since there are two different masked input tokens, there is an almost certain probability that some tokens will be simultaneously double-masked. As a remedy, we propose that for a token that is masked in both instances, the recovered feature should be as close as possible in both cases. This naturally introduces an explicit constraint at the feature level, ensuring that the encoder outputs more informative features for downstream tasks. As a result, Fig.~\ref{fig:contras_fig} shows that the proposed Point-CMAE can perform a decent convergence with obvious improvement at the beginning of the fine-tuning compared to the contrastive counterpart.

In summary, our main contributions are recapped as follows:
\begin{enumerate}[nosep]
    \item We experimentally validate that integrating contrastive learning into the MAE paradigm for point cloud pre-training with ViTs architecture is non-trivial and often leads to a severe performance drop.

    \item To address the aforementioned issue, we propose Point-CMAE, a method that enhances the MAE paradigm by integrating the advantages of contrastive learning through a straightforward yet effective design. This significantly enhances the informativeness and representativeness of the encoder.
    
    \item Extensive experimental results on various point cloud downstream tasks such as object classification, part segmentation, and few-shot classification demonstrate that our proposed Point-CMAE achieves new state-of-the-art performance under the standard ViTs setting for single-modal data.
\end{enumerate}

%% file: sections/2_relatedworks.tex
\section{Related Work: Self-supervised Learning for Point Cloud}
\label{sec:related-work}
Self-supervised learning (SSL), as one type of unsupervised learning where the supervision signals can be acquired from the data itself, has attracted more and more attention to computer vision~\cite{grill2020bootstrap,he2022masked}. 
As a result, a lot of methods have been proposed to advance this technique from different perspectives~\cite{bardes2021vicreg,chen2020simple,grill2020bootstrap,he2022masked,tang2023deep,oord2018representation, zbontar2021barlow,ren2024survey}. 
Especially, recent 3D point cloud understanding also embraces a promising development owing to the SSL. Similarly to the image domain, these approaches can be mainly divided into two mainstream, \ie, the contrastive pre-training and the generative pre-training approaches.

\noindent \textbf{Contrastive Pre-training} based approaches~\cite{rao2020global, chen2021exploring, xie2020pointcontrast, huang2021spatio} aim to learn instance discriminative representations to distinguish one sample from the others. 
Especially, PointContrast~\cite{xie2020pointcontrast}, as the pioneering approach that constructs two point clouds from different perspectives and compares point feature similarities for point cloud pre-training.
Info3D~\cite{sanghi2020info3d} aims to maximize the mutual information between the 3D shape and a geometrically transformed version of the same shape with a MoCo~\cite{he2020momentum}-like memory bank for caching the negative examples. 
ProposalContrast~\cite{yin2022proposalcontrast} enhances proposal representations by analyzing the geometric point relationships within each proposal, achieving this by optimizing for inter-cluster and inter-proposal separation to better adapt to 3D detection properties. 
FAC~\cite{liu2023fac} forms advantageous point pairs from the same foreground segment with similar semantics and captures feature correlations within and across different point cloud views using adaptive learning. 
Note that the above methods that explored the contrastive training for point cloud are all based on CNN backbones. With the overwhelming development of ViTs, Point-BERT~\cite{PointBERT} firstly includes the classic MoCo-style contrastive learning into ViTs-based pre-training pipeline, but as a side exploration compared to its generative learning property.
MaskPoint~\cite{MaskPoint} converts the point cloud into discrete occupancy values, using binary classification to distinguish masked object points and sampled noise points. 
SoftClu~\cite{mei2022data} and CluRender~\cite{mei2024unsupervised} use clustering and rendering for point-level supervision, extracting discriminative features without data augmentation.
More recently, ReCon~\cite{qi2023contrast} integrated contrastive learning into pre-training to enhance performance within its generative pipeline. However, their approach employs supervised contrastive learning with a pre-constructed label set to mitigate overfitting, a prevalent concern in ViTs-based methods. Nonetheless, developing a self-supervised contrastive learning strategy tailored for ViTs-based point cloud pre-training remains challenging, particularly in effectively addressing the overfitting issue, which poses a significant obstacle in this domain.

\noindent \textbf{Generative Pre-training} based approaches are proposed inspired by the success of \textit{mask and reconstruct} strategy used in BERT from NLP~\cite{devlin2018bert} to MAE~\cite{he2022masked} in vision with Transformers~\cite{vaswani2017attention,dosovitskiy2020image,ren2023masked}. It prioritizes the encoder's pre-training by reconstructing masked information or its 2D projections. Notable examples include Point-BERT~\cite{PointBERT} and Point-MAE~\cite{MaskPoint}, which are proposed for point cloud pre-training with the masked reconstruction strategy. 
Point-M2AE~\cite{zhang2022point} develops a hierarchical network that effectively models geometric and feature information progressively. 
The up-following works like TAP~\cite{wang2023take} and Ponder~\cite{huang2023ponder} focus on generating 2D projections of the point cloud as part of their pre-training strategies. 
Joint-MAE~\cite{guo2023joint} addresses the correlation between 2D images and 3D point clouds, introducing hierarchical modules for cross-modal interaction to reconstruct masked information across both modalities.
PointGPT~\cite{chen2024pointgpt} extends the concept of GPT~\cite{radford2018improving} to point clouds pre-training with post-pre-training with larger datasets.
PointDif~\cite{zheng2023point} concentrates on refining the training approach with diffusion models~\cite{ho2020denoising}. 
Despite MAE models exhibiting favorable optimization
properties~\cite{wei2022contrastive} and delivering promising performance, their focus is on learning relationships among the tokens within the same input image, rather than modeling the relation among different samples as in contrastive learning, which results in less discriminative learned representations~\cite{huang2023contrastive} or data filling issues~\cite{qi2023contrast,xie2023data}. 
Moreover, in the context of point cloud reconstruction, the commonly used chamfer loss compels the model to precisely match the ground truth set. However, the ground truth itself represents just one sample from the true underlying distribution, posing a challenging optimization problem that often results in suboptimal performance~\cite{MaskPoint}. To tackle this issue, we propose injecting feature-level contrastive properties into MAE pre-training.

%% file: sections/3_methods.tex
\section{Methodology}
\label{sec:method}
\begin{figure}[!t]
    \centering
    \includegraphics[width=1.0\linewidth]{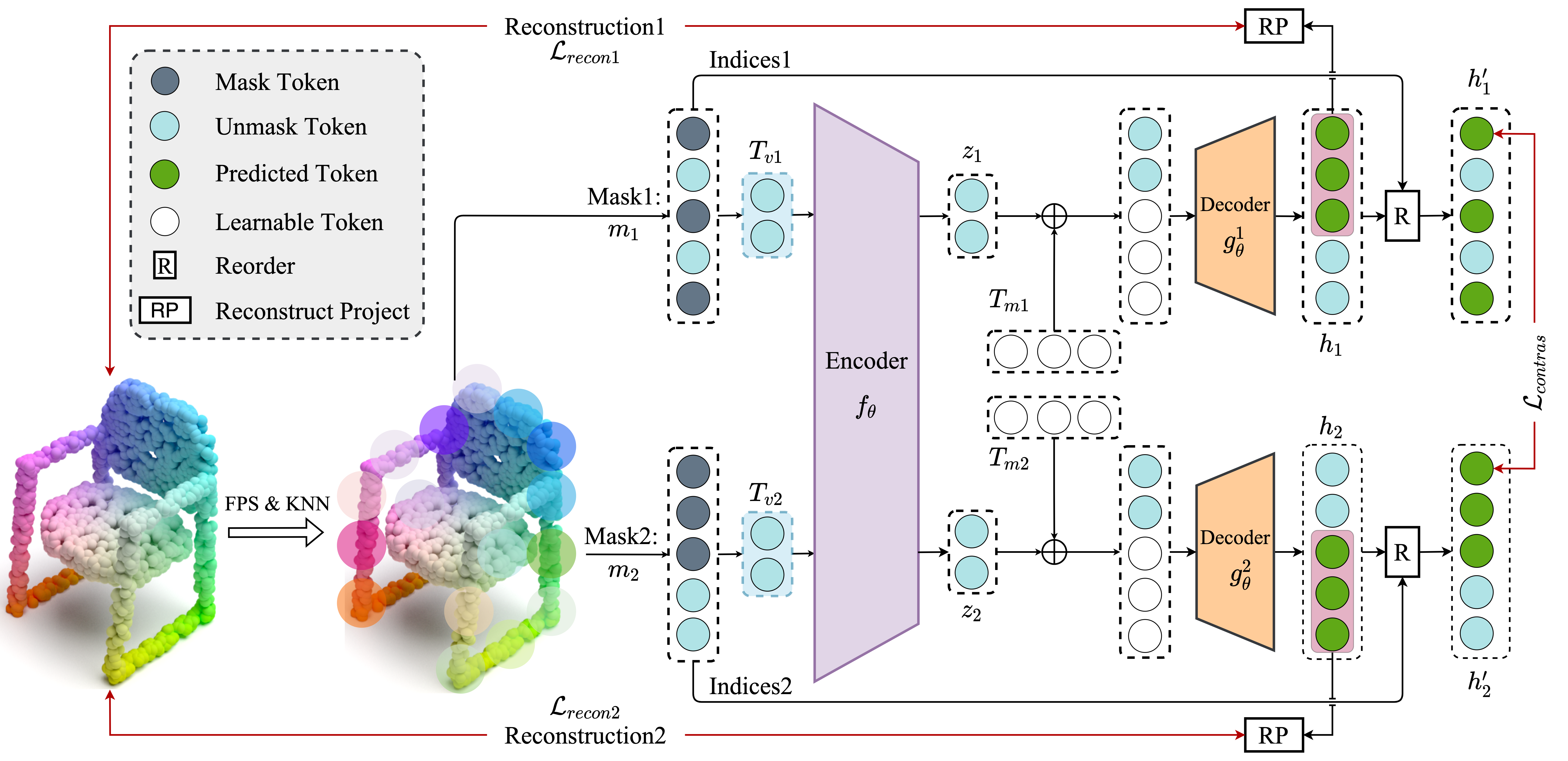}
    \caption{The framework of the proposed Point-CMAE. The symbols $\oplus$ denote the token dimension concatenation. The point patch embedding is denoted as "FPS\&KNN". The symbol $\oplus$ denotes the token dimension concatenation.}
    \vspace{-2em}
    \label{fig:framework}
\end{figure}
The overview of the proposed method Point-CMAE is illustrated in Fig.~\ref{fig:framework}. Before diving into the detailed introduction of the proposed Point-CMAE, we first provide the point cloud embedding and the masked autoencoder for point cloud pre-training by ViTs in Sec.~\ref{subsec:pre}. Then the explanation of why the proposed method is designed via dual-masking augmentation is introduced in Sec.~\ref{subsec:dual}. Building upon the proposed dual-masking framework, we explicitly introduce contrastive properties into the MAE paradigm, as detailed in Sec~\ref{subsec:explicit_contras}. 

\subsection{Preliminaries}
\label{subsec:pre}
\noindent{\textbf{Point Patch Embedding.}}
Different from images that lie on regular grids which can be naturally divided into patches, point clouds are known to be irregular and less
structured, based on this property, we follow Point-BERT~\cite{PointBERT} first to divide the input point cloud ($X^{i} \in \mathbb{R}^{p \times 3}$, $p$ denotes the number of the points) into irregular point patches via Farthest Point Sampling (\ie, $\operatorname{FPS(\cdot)}$) and K-Nearest Neighborhood (\ie, $\operatorname{KNN(\cdot)}$) algorithm and output $n$ center points $c$ ($c = \operatorname{FPS}(X^{i}), c \in \mathbb{R}^{n \times 3}$) and the corresponding neighbor points $P$ ($P = \operatorname{KNN}(X^{i}, c), P \in \mathbb{R}^{n \times k \times 3}$) of each center point. Finally, a lightweight PointNet~\cite{PointNet} (\ie, $\operatorname{PointNet(\cdot)}$) which mainly consists of MLPs is applied to the point patches (usually only for the visible point patches) to achieve the embedded tokens $T$ ($T = \operatorname{PointNet}(P), T \in \mathbb{R}^{n \times C}$, $C$ denotes the embedding dimension).

\noindent{\textbf{Masked Autoencoder for point cloud with ViTs.}} 
For the embedded tokens $T$, to deploy the MAE strategy requires a mask $m$ applied to $T$ and outputs the visible tokens $T_{v} \in \mathbb{R}^{(1-r)n \times C}$ and the masked tokens $T_{m} \in \mathbb{R}^{rn \times C}$, where $r$ denotes the mask ratio. Then the MAE of the point cloud~\cite{PointBERT,PointMAE} can be summarized as point tokens which are masked with random mask $m$ are fed into the encoder $f_{\theta}(\cdot)$, and then the decoder $g_{\phi}(\cdot)$ predicts the original masked points $X_m$ with distribution $\mathcal{D}$:
\begin{equation}
    \min_{\theta, \phi} \underset{\substack{X \sim \mathcal{D}}}{\mathbb{E}} \left[\mathcal{M} \left( g_{\phi}(z \oplus T_m), X_{m}\right)\right], \quad z = f_{\theta}(T_{v}).
    \label{eq:eq1}
\end{equation}
Here $z \in \mathbb{R}^{rn \times C}$ denotes the latent feature of visible tokens $T_{v}$, $\mathcal{M}$ denotes the similarity measurement, and it was usually set as the Chamfer-Distance (\ie, $\operatorname{CD(\cdot)}$)~\cite{fan2017point}. The symbol $\oplus$ denotes the token dimension concatenation. $\theta$ and $\phi$ are the trainable parameters of the encoder and the decoder, respectively.

\subsection{Better Initialization for ViTs Encoder with Dual Masking}
\label{subsec:dual}
Inspired by the conclusion drawn in ~\cite{kong2023understanding} that \textit{the output feature is robust only if the most significant input part is not masked out}. Thereby, we propose to increase the invariance property of ViTs pre-training by increasing the probability of the most important input point patch that will not be masked. 
Unlike the heavy augmentation strategies (\eg, color jittering) adopted in the image, point clouds contain only the position information. Moreover, point cloud pre-training is very sensitive to the geometry data augmentation (\ie, a simple rotation usually brings a large performance increase)~\cite{dong2022autoencoders,qi2023contrast}, which will lead to ambiguity in knowing whether the contribution is made by the geometry data augmentation or the method itself. 

To this end, we propose using two masks, $m_1$ and $m_2$, that share the same mask ratio $r$ but incorporate different sources of randomness (\ie, $m_1 \ne m_2$) as the augmentation operation. The same encoder $f_\theta$ is then used to process the visible tokens $T_{v_1}$ and $T_{v_2}$, outputting the encoded features $z_1$ and $z_2$. Next, two separate decoders, $g_{\theta}^{1}$ and $g_{\theta}^{2}$, which share the same architecture but have weights that are updated differently based on their own inputs, are used. The main idea of the proposed method is illustrated in Fig.~\ref{fig:framework}. Finally, the output of each decoder is projected back to the point cloud space for two separate reconstruction chamfer losses. Based on Eq.~\ref{eq:eq1}, the loss function can now be rewritten as follows:
\begin{equation}
\begin{aligned}
\mathcal{L}_{re} &= \mathcal{L}_{recon1} + \mathcal{L}_{recon2} \\ &=
    \operatorname{CD}(\operatorname{RP}(g_{\phi}^{1}(z_1)), X_{m_1}) + \operatorname{CD}(\operatorname{RP}(g_{\phi}^{2}(z_2)), X_{m_2}),
    \label{eq:eq2}
\end{aligned}
\end{equation}
here $\operatorname{RP(\cdot)}$ is a fully connected layers-based projection head aiming to reconstruct masked point patches via projecting only the masked features back to the point coordinate cloud space. 
This simple yet effective design brings three advantages for point cloud pre-training:
i) It decreases the possibility that a significant input part will be masked out (\eg, though this token will be masked via $m_1$, now we have the possibility that it will not be masked out in $m_2$).
ii) Two separate decoders force the encoder to learn better representations since two decoders for two different masks $m_1$ and $m_2$ require the encoder's output to be more informative to meet the requirements of both $g_{\phi}^{1}$ and $g_{\phi}^{2}$.
iii) It naturally enriches the MAE paradigm with contrastive properties that largely improve the classification performance.
Both i) and ii) are supported by the experimental results in Sec.~\ref{sec:experiments} while the analysis of iii) is provided in Sec.~\ref{subsec:explicit_contras}.

\subsection{Explicit Feature Level Contrastive Constraint}
\label{subsec:explicit_contras}
Given the ground truth points for the point cloud that are also one of the samplings of the original point cloud, directing using the Chamfer loss to minimize the difference between the ground truth points and the predicted points usually leads to a sub-optimize issue~\cite{MaskPoint}, especially for point cloud data that contains the position of each point that is largely different from the MAE in image domain where the pixel-level contraction is naturally more informative to reconstruct.

With the same masking ratio \(r\) for masks \(m_1\) and \(m_2\) in the dual-masking pipeline, there is a substantial probability that a token in the embedded tokens \(T \in \mathbb{R}^{n \times C}\) could be masked simultaneously by both masks. The probability of this occurring can be calculated as:
\begin{equation}
\begin{aligned}
    p = 1 - (1-r^{2})^{n}.
    \label{eq:eq3}
\end{aligned}
\end{equation}
In particular, given an example that $r=0.6$, $n=64$, $p \approx 0.945 >> 0$. Based on this observation, we propose to let the features $h_1$ and $h_2$ from both the decoders $g_{\phi}^{1}$ and $g_{\phi}^{2}$ of a certain point token that can be as close as possible to the feature level. Specifically, because in point-cloud MAE pretraining, the visible token after the encoder $f_\theta$ is directly concatenated with its corresponding masked token before passing through the corresponding decoder, and as the results, the output of the decoder still follows the same [visible, masked] order. This makes it non-trivial to find a token that was masked by both $m_1$ and $m_2$. As a remedy, we first recorded the output features from both decoders based on the indices when conducting the masking operations before $f_{\theta}$:
\begin{equation}
\begin{aligned}
    h^{\prime}_{1} = \operatorname{R}(Indices1, h_{1}),\\ h^{\prime}_{2} = \operatorname{R}(Indices2, h_{2}),
\end{aligned}
\end{equation}
here $\operatorname{R(\cdot)}$ indicates reorder operation. Then the contrastive constraint can be written as:
\begin{equation}
\begin{aligned}
    \mathcal{L}_{contras} = {\frac{m_1 \cap m_2}{|m_1 \cap m_2|}}\sum_{i}^{n}(1 - \mathcal{M}(h^{\prime}_{1i}, h^{\prime}_{2i})),
    \label{eq:eq4}
\end{aligned}
\end{equation}
here $|m_1 \cap m_2|$ denotes the number of the co-masked point tokens by both $m_1$ and $m_2$. $\mathcal{M}$ is the cosine similarity measurement. Then the total optimization objective can be written as follows:
\begin{equation}
    \mathcal{L} = \mathcal{L}_{re} + \lambda \mathcal{L}_{contras},
\end{equation}
$\mathcal{L}_{contras}$ here serve as a regularization term of $\mathcal{L}_{re}$, and $\lambda$ is the regularization weight. As a result, the feature-level regularization, $\mathcal{L}_{contras}$, brings MAE explicit contrastive properties to ease the sub-optimize issue that was inherited from the MAE-based point cloud pre-training induced by the Chamfer loss~\cite{MaskPoint}.

%% file: sections/4_experiments.tex
\section{Experiments}
\label{sec:experiments}
\subsection{Self-supervised Pre-training Setups}
\noindent{\textbf{Pre-training.}} We pre-train the proposed method ShapeNet~\cite{chang2015shapenet}. ShapeNet is a synthetic 3D dataset that contains 52,470 3D shapes across 55 object categories. 
We pre-train our model only on the training set, which contains 41,952 shapes. 
For each 3D shape, we sample 1024 points to serve as the input for the mode, We set $n$ as 64, which means each point cloud is divided into 64 patches. 
Furthermore, the KNN algorithm selects the $k=32$ nearest point as a point patch. Following ~\cite{MaskPoint,PointMAE}, the proposed method is pre-trained for 300 epochs using an AdamW optimizer~\cite{loshchilov2018decoupled}.
In the autoencoder’s backbone, the encoder has 12 Transformer blocks while the decoder has 4 ViTs encoder blocks. 
Each Transformer block has 384 hidden dimensions and 6 heads. MLP ratio in Transformer blocks is set to 4.
The batch size was set to 128 during the entire pre-training.
The initial learning rate was set to 0.0005 with cosine learning rate decay (the decay weight was 0.05) employed. 
More details regarding our experimental configuration and implementation are provided in our supplementary materials (\ie, \textit{Supp. Mat.}) 

\input{tables/scanobjectnn_results}
\noindent{\textbf{Transfer Protocol.}} 
Similar to ~\cite{dong2022autoencoders,qi2023contrast}, we adopt three variants of transfer learning protocols for classification tasks during fine-tuning. \ie, 
(a) \textbf{Full}: Fine-tuning pre-trained models by updating all backbone and classification heads. 
(b) \textbf{MLP-Linear}: The classification head is a single-layer linear MLP, and we only update these head parameters during fine-tuning.
(c) \textbf{MLP-3}: The classification head is a three-layer non-linear MLP (\ie, the same as the
one used in FULL), and we only update these head parameters during fine-tuning.

\subsection{Transfer Learning on Downstream Tasks}
To assess the efficacy of the pre-trained model, we gauged its performance on various fine-tuned tasks using numerous real-world datasets.

\noindent{\textbf{3D Real-Word Object Classification.}}
We use the scanned ScanObjectNN~\cite{uy2019revisiting} dataset to evaluate the shape recognition ability of the pre-trained model of our method. The ScanObjectNN~\cite{uy2019revisiting} dataset covers around 15K real-world objects from 15 categories, and it is divided into three subsets: OBJ-BG (objects and background), OBJ-ONLY (only objects), and PB-T50-RS (objects, background, and artificially added perturbations). For a fair comparison, we report the results without voting strategy~\cite{liu2019relation}. 
The results shown in Tab.~\ref{tab:scanobjectnn} indicate that: 
(i) Without increasing the parameters, the proposed Point-CMAE archives obvious accuracy increase with standard ViTs architecture under the single-modal full-tuning protocol and even outperforms the supervised counterparts. There is an absolute by \textbf{0.77\%} improvement even under the most challenging PB\_T50\_RS setting.
(ii) When using the commonly used rotation data augmentation during fine-tuning as ACT or PointGPT, the proposed Point-CMAE can still outperform ACT on OBJ\_BG (by \textbf{0.17\%}) and PB\_T50\_RS (by \textbf{0.54\%}) even both ACT and PointGPT used the multi-modal information.
(iii) The performance of our Point-CMAE on MLP-Linear and MLP-3 further demonstrates that the generalization ability of the pre-trained model is improved even when only fine-tuning the classification head.

\input{tables/modelnet_two_col}
\noindent{\textbf{3D Synthetic Object Classification.}}
We also show the evaluation of 3D shape classification on the synthetic dataset ModelNet40~\cite{wu20153d}. ModelNet40 consists of 12,311 clean 3D CAD models, covering 40 object categories. We follow standard protocols as ~\cite{PointBERT,PointMAE} to split ModelNet40 into 9,843 instances for the training set and 2,468
for the testing set. Standard random scaling and random translation are applied for data augmentation during training. The results are shown in Tab.~\ref{tab:modelnet}. 
Note that for a fair comparison, the voting strategy was adopted for reproducing Point-MAE~\cite{PointMAE} and the proposed Point-CMAE under the Full evaluation protocol. Tab.~\ref{tab:modelnet} shows that: 
(i) The proposed Point-CMAE also works effectively on the synthetic dataset and with a slight improvement compared to our baseline methods. 
(ii) Such effectiveness is further validated under the MLP-Linear and MLP-3 protocols where the pre-trained backbone is frozen.

\input{tables/fewshot_wrap}
\input{tables/partseg_shapenetpart}
\noindent{\textbf{Few-shot Classification.}}
We follow previous works~\cite{PointMAE,PointBERT,MaskPoint,qi2023contrast} to conduct few-shot learning experiments on ModelNet40~\cite{wu20153d}, adopting $num\_cls$-way, $num\_sample$-shot setting, where $num\_cls$ is the number of classes that randomly selected from the dataset and $num\_sample$ is the number of objects randomly sampled for each class. 
We use the above-mentioned $num\_cls \times num\_sample$ objects for training. 
During testing, we randomly sample 20 unseen objects
from each of $num\_cls$ classes for evaluation. 
The results with the setting of $num\_cls$ in {5, 10} and $num\_sample$ in {10, 20} are presented in Tab.~\ref{tab:few-shot}. Following all three protocols, we conduct 10 independent experiments for each setting and report mean accuracy with standard deviation. Tab.~\ref{tab:few-shot} shows that:
(i) Besides the truth that our self-supervised Point-CMAE outperforms the supervised solutions by a large margin, Point-CMAE brings significant improvements of +8.8\%,
+4.7\%, +8.1\%, +5.9\% respectively for the four
settings over from scratch FULL transferring
baseline~\cite{vaswani2017attention}.
(ii) The proposed Point-CMAE not only consistently achieves
the best performance compared to our baseline method Point-MAE~\cite{PointMAE}, but it is also worth pointing out that with the standard ViTs architecture pre-trained with only the single point cloud data, Point-CMAE achieves competitive or even better performance compared to other state-of-the-art methods which were proposed either with complex hierarchical ViTs structure (\ie, Point-M2AE) or trained with multi-modal information (\ie, Joint-MAE, Point-GPT, and ACT).
(iii) For the MLP-Linear and MLP-3 transfer protocol, we observed that the Point-CMAE achieves obvious improvements (\eg, 2.4\%, 5.7\%, and 2.6\% for the 5way20shot, 10way10shot, and the 10way20shot under the MLP-Linear protocol) compared to Point-MAE with smaller deviations.

\noindent{\textbf{3D Part Segmentation.}}
We evaluate the segmentation performance of the proposed Point-CMAE on ShapeNetPart~\cite{yi2016scalable} dataset. It contains 16,881 objects of 2,048 points from 16 categories with 50 parts in total. The segmentation head in our method is the same as in Point-MAE~\cite{PointMAE}, which is relatively simple and does not use any propagating operation or DGCNN~\cite{DGCNN}. We use the learned features from the 4th, 8th, and 12th layers of the Transformer block, and concatenate the three levels of features. 
Then an average pooling and a max pooling were applied separately to obtain two global features. For Transformer, we used per-category mean IoU ($\mbox{mIoU}_C$) and mean IoU averaged over all test instances ($\mbox{mIoU}_I$) to assess performance. The part segmentation results are provided in Tab.~\ref{tab:partseg}, which shows that:
(i) The proposed Point-CMAE archives the best per-category mean IoU and the second-best mean IoU averaged over all test instances. Especially, a significant improvement 0.7\% compared to Point-MAE~\cite{PointMAE} on $\mbox{mIoU}_C$.
(ii) When compared to other state-of-the-art methods that adopt hierarchical ViTs architecture (\ie, Point-M2AE), utilize post-training on extra-large datasets (\ie, PointGPT-L), or incorporate multi-modal information (\ie, ACT and Recon), the proposed Point-CMAE still slightly outperforms these methods in terms of the evaluation metric $\mbox{mIoU}_C$.

\begin{figure}[!t]
    \centering
    \includegraphics[width=1.0\linewidth]{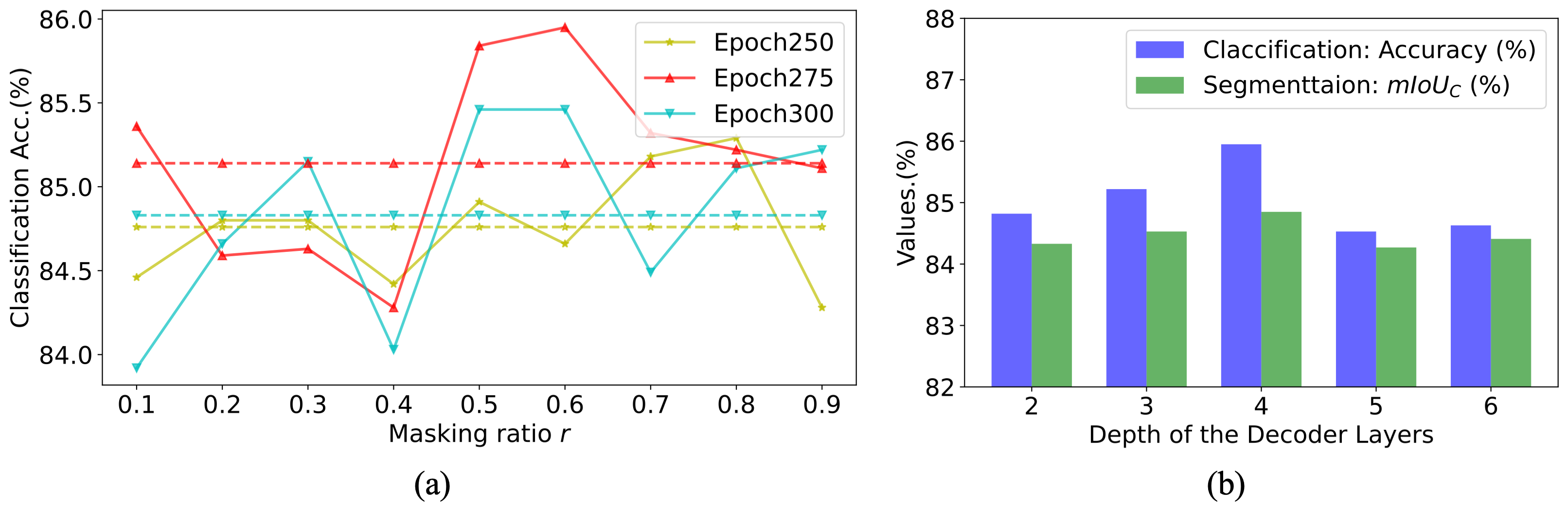}
    \vspace{-5mm}
    \caption{(a) The classification results on the ScanobjectNN~\cite{uy2019revisiting} (PB\_T50\_RS) dataset cross different masking ratios. The corresponding average results across the entire masking ratios are depicted with the dashed lines. (b) Both the classification on the ScanobjectNN~\cite{uy2019revisiting} (PB\_T50\_RS) and the part segmentation results on the ShapeNetPart~\cite{yi2016scalable} are provided to study how the depth of the decoder affects the pre-traing.}
    \vspace{-5mm}
    \label{fig:ablation}
\end{figure}

\subsection{Ablation Study}
\noindent{\textbf{The sweet spot of the mask ratio.}} The mask ratio has been validated significantly for MAE-based self-supervised learning in both the image and point cloud domains~\cite{he2022masked,PointBERT,PointMAE}. To determine a suitable masking ratio for our method, we varied the mask ratio from 0.1 to 0.9. The corresponding results are depicted in Fig.~\ref{fig:ablation} (a). Specifically, using a pre-trained Point-CMAE model, we followed Recon~\cite{qi2023contrast} and evaluated the model checkpoints from epochs 250, 275, and 300. We found that the checkpoint from epoch 275 performs well for our method, as indicated by the average results across mask ratios shown with dashed lines in Fig.~\ref{fig:ablation} (a). A mask ratio of 0.6 consistently produced the best results compared to other ratios. Therefore, we adopted a mask ratio of 0.6 throughout this paper, which is also consistent with findings in related works~\cite{PointMAE,qi2023contrast}.

\noindent{\textbf{The effect of the depth of the ViTs decoder.}} Fig.~\ref{fig:ablation} (b) shows the fine-tuning performance for both the classification and the part segmentation tasks with different numbers of the ViTs layers within the decoder.
It can be seen that the performance increase when the depth is in increased from 2 to 4, while it decrease when the depth is further increased from 4 to 6. We set the decoder 4 ViTs transformer layers throughout our work based on this observation. This setting is also consistent with our baseline method Point-MAE~\cite{PointMAE}.

\input{tables/ablation}
\noindent{\textbf{The effect of each component.}}
The ablation studies regarding how each component affects both the classification and the part segmentation performance of the proposed Point-CMAE are shown in Tab.~\ref{tab:ablation}. It mainly uncovers that:
(i) When just using the proposed dual-masking strategy (Sec.~\ref{subsec:dual}) from (a), an obvious improvement can be achieved for both the classification and the segmentation.
(ii) Using one same encoder during pre-training is better than using two separate encoders.
(iii) Using the same decoder to reconstruct the point cloud from two masks degenerates the overall performance for both the classification and the segmentation tasks. 
(iv) The contrastive learning strategy proposed in our paper brings obvious improvements. In particular, built upon the same encoder and two separate decoders during pre-training, (f) archives the best results compared to the rest, and we set (f) as our full model.

%% file: tables/scanobjectnn_results.tex
\begin{table}[t!]
    \centering
    \setlength\tabcolsep{7.5pt}
    \setlength{\extrarowheight}{1.5pt}
    \caption{Classification results on ScanObjectNN.  \texttt{DA}: rotation data augmentation is used during fine-tuning.
    The overall accuracy, \ie, OA (\%) is reported. 
    }
    \label{tab:scanobjectnn}
    \resizebox{\linewidth}{!}{
    \begin{threeparttable}
    \begin{tabular}{lccccc}
    \toprule[0.95pt]
    Method & \#Params(M) & \texttt{DA} & OBJ\_BG & OBJ\_ONLY & PB\_T50\_RS\\
    \midrule[0.6pt]
    \multicolumn{6}{c}{\textit{Supervised Learning Only}}\\
    \midrule[0.6pt]
    PointNet~\cite{PointNet} & 3.5  & - & 73.3 & 79.2 & 68.0\\
    SpiderCNN~\cite{SpiderCNN} & -  & - & 77.1 & 79.5 & 73.7\\
    PointNet++~\cite{PointNet++} & 1.5  & - & 82.3 & 84.3 & 77.9\\
    DGCNN~\cite{DGCNN} & 1.8 & - & 82.8 & 86.2 & 78.1\\
    PointCNN~\cite{PointCNN} & 0.6 & - & 86.1 & 85.5 & 78.5\\
    BGA-DGCNN~\cite{uy2019revisiting} & 1.8 & - & - & - & 79.7\\
    BGA-PN++~\cite{uy2019revisiting} & 1.5 & - & - & - & 80.2\\
    DRNet~\cite{qiu2021dense} & - & - & - & - & 80.3\\
    GBNet~\cite{qiu2021geometric} & 8.8 & - & - & - & 80.5\\
    SimpleView~\cite{goyal2021revisiting} & - & - & - & - & 80.5$\pm$0.3\\
    PRANet~\cite{cheng2021net} & 2.3 & -& - & - & 81.0\\
    MVTN~\cite{hamdi2021mvtn} & - & - & - & - &82.8\\
    PointMLP~\cite{ma2021rethinking} & 13.2 & - & - & - & 85.4$\pm$0.3\\
    \midrule[0.6pt]
    \multicolumn{6}{c}{\textit{with Standard ViTs and Single-Modal Self-Supervised Learning} ({\scshape Full})}\\
    \midrule[0.6pt]
    Transformer~\cite{vaswani2017attention} & 22.1 & $\times$ & 79.86 & 80.55 & 77.24\\
    OcCo~\cite{wang2021unsupervised} & 22.1 & $\times$ & 84.85 & 85.54 & 78.79\\
    Point-BERT~\cite{PointBERT} & 22.1 & $\times$ & 87.43 & 88.12 & 83.07\\
    MaskPoint~\cite{MaskPoint} & 22.1 & $\times$ & \underline{89.30} & 88.10 & 84.30\\
    Point-MAE~\cite{PointMAE} & 22.1 & $\times$ & \textbf{90.02} & \underline{88.29} & \underline{85.18} \\
    \rowcolor{linecolor1}Point-CMAE (Ours) & 22.1 & $\times$ & \textbf{90.02} & \textbf{88.64} & \textbf{85.95}\\
    \rowcolor{linecolor}Point-CMAE (Ours) & 22.1 & $\checkmark$ & \textbf{93.46} & \textbf{91.05} & \textbf{\textbf{88.75}}\\
    \midrule[0.6pt]
    \multicolumn{6}{c}{\textit{with Hierarchical ViTs / Multi-Modal/Post-Process Self-Supervised Learning} ({\scshape Full})} \\
    \midrule[0.6pt]
    Point-M2AE~\cite{zhang2022point} & 15.3 & $\times$  & 91.22 & 88.81 & 86.43\\
    Joint-MAE~\cite{guo2023joint} & - & $\times$ & 90.94 & 88.86 & 86.07 \\
    ACT~\cite{dong2022autoencoders} & 22.1 & $\checkmark$  & 93.29 & 91.91 & 88.21\\
    PointGPT-S~\cite{chen2024pointgpt} & 29.2 & $\checkmark$ & 93.39 & 92.43 & 89.17 \\
    \midrule[0.1pt]
    \midrule[0.1pt]
    \multicolumn{6}{c}{\textit{with Standard ViTs and Single-Modal Self-Supervised Learning} ({\scshape MLP-Linear})} \\
    \midrule[0.6pt]
    Point-MAE~\cite{PointMAE} & 22.1 & $\times$ & 82.58 $\pm$ 0.58 & \textbf{83.52} $\pm$ {0.41} & 73.08 $\pm$ 0.30 \\
    \rowcolor{linecolor}Point-CMAE (Ours) & 22.1 & $\times$ & \textbf{83.48 $\pm$ 0.31} & 83.45 $\pm$ \textbf{0.35} & \textbf{73.15 $\pm$ 0.11} \\
    \midrule[0.6pt]
    \multicolumn{6}{c}{\textit{with Standard ViTs and Single-Modal Self-Supervised Learning} ({\scshape MLP-3})} \\
    \midrule[0.6pt]
    Point-MAE~\cite{PointMAE} & 22.1 & $\times$ & 84.29 $\pm$ 0.55 & 85.24 $\pm$ 0.41 & 77.34 $\pm$ \textbf{0.12} \\
    \rowcolor{linecolor}Point-CMAE (Ours) & 22.1 & $\times$ & \textbf{85.88 $\pm$ 0.53} & \textbf{85.60 $\pm$ 0.35} & \textbf{77.47} $\pm$ {0.13} \\
    \bottomrule[0.95pt]
    \end{tabular}
    \end{threeparttable}
    }
\vspace{-5mm}
\end{table}

%% file: tables/modelnet_two_col.tex
\begin{table}[t!]
      \caption{Classification results on the ModelNet40 dataset. The overall accuracy, \ie, OA (\%) is reported. \texttt{[ST]}: standard Transformer architecture. $^{*}$: The reproduced results.}
      \label{tab:modelnet}
 \begin{minipage}[t]{0.48\textwidth}
    \centering
    \resizebox{\linewidth}{!}{
    \setlength\tabcolsep{2.8pt}
    \begin{tabular}{lccc}
    \toprule[0.81pt]
    Method & \texttt{[ST]} & \#Point & OA (\%)\\
    \midrule[0.6pt]
    \multicolumn{4}{c}{\textit{Supervised Learning Only}}\\
    \midrule[0.6pt]
    PointNet~\cite{PointNet} & - & 1k P & 89.2\\
    PointNet++~\cite{PointNet++} & - & 1k P & 90.7\\
    PointCNN~\cite{PointCNN} & - & 1k P & 92.5\\
    DGCNN~\cite{DGCNN} & - & 1k P & 92.9\\
    DensePoint~\cite{liu2019densepoint} & - & 1k P & 93.2\\
    PointASNL~\cite{wu2019pointconv} & - & 1k P & 92.9\\
    DRNet~\cite{qiu2021dense} & - & 1k P & 93.1\\
    \midrule[0.6pt]
    Point Trans.~\cite{engel2021point} & $\times$ & 1k P & 92.8\\
    PCT~\cite{guo2021pct} & $\times$ & 1k P & 93.2\\
    PointTransformer~\cite{zhao2021point} & $\times$ & 1k P & 93.7\\
    NPCT~\cite{guo2021pct} & $\checkmark$ & 1k P & 91.0\\
    \bottomrule[0.81pt]
    \end{tabular}
    }
	
  \end{minipage}
  \hspace{3pt}
  \begin{minipage}[t]{0.48\textwidth}
   \centering
        \makeatletter\def\@captype{table}\makeatother
	\centering
	\setlength\tabcolsep{3.2pt}
        \setlength{\extrarowheight}{1.4pt}
	\resizebox{\linewidth}{!}{
    \begin{tabular}{lccc}
    \toprule[0.99pt]
    Method & \texttt{[ST]} & \#Point & OA (\%)\\
    \midrule[0.6pt]
    \multicolumn{4}{c}{\textit{with Self-Supervised Representation Learning} ({\scshape Full})}\\
    \midrule[0.6pt]
    Transformer~\cite{vaswani2017attention} & $\checkmark$ & 1k P & 91.4\\
    OcCo~\cite{wang2021unsupervised} & $\checkmark$ & 1k P & 92.1\\
    Point-BERT~\cite{PointBERT} & $\checkmark$ & 1k P & 93.2\\
    Point-MAE~\cite{PointMAE} & $\checkmark$ & 1k P & 93.8\\
    Point-MAE$^{*}$~\cite{PointMAE} & $\checkmark$ & 1k P &\underline{93.5}\\
    \rowcolor{linecolor}Point-CMAE (Ours) & $\checkmark$ & 1k P & \textbf{93.6}\\
    \midrule[0.6pt]
    \multicolumn{4}{c}{\textit{with Self-Supervised Representation Learning} ({\scshape Mlp-Linear})}\\
    \midrule[0.6pt]
    Point-MAE~\cite{PointMAE} & $\checkmark$ & 1k P & 91.22$\pm$0.26\\
    \rowcolor{linecolor}Point-CMAE (Ours) & $\checkmark$ & 1k P & \textbf{92.30}$\pm$\textbf{0.32}\\
    \midrule[0.6pt]
    \multicolumn{4}{c}{\textit{with Self-Supervised Representation Learning} ({\scshape Mlp-$3$})}\\
    \midrule[0.6pt]
    Point-MAE~\cite{PointMAE} & $\checkmark$ & 1k P & 92.33$\pm$0.09\\
    \rowcolor{linecolor}Point-CMAE (Ours) & $\checkmark$ & 1k P & \textbf{92.60}$\pm$\textbf{0.19} \\
    \bottomrule[0.99pt]
    \end{tabular}
    }
   \end{minipage}
\vspace{-5mm}
\end{table}

%% file: tables/fewshot_wrap.tex
\begin{table}[t!]
    \centering
    \caption{Few-shot classification on \textbf{ModelNet40}, overall accuracy (\%) is reported.}
    \label{tab:few-shot}
    \setlength\tabcolsep{5pt}
    \setlength{\extrarowheight}{1pt}
    \scalebox{0.97}{
    \begin{tabular}{lcccc}
    \toprule[0.95pt]
    \multirow{2}{*}[-0.5ex]{Method}& \multicolumn{2}{c}{\textbf{5-way}} & \multicolumn{2}{c}{\textbf{10-way}}\\
    \cmidrule(lr){2-3}\cmidrule(lr){4-5}
    & 10-shot & 20-shot & 10-shot & 20-shot\\
    \midrule[0.6pt]
    \multicolumn{5}{c}{\textit{Supervised Representation Learning}}\\
    \midrule[0.6pt]
    PointNet~\cite{PointNet} &52.0 $\pm$ 3.8 &  57.8 $\pm$ 4.9&  46.6 $\pm$  4.3& 35.2 $\pm$ 4.8 \\
    PointNet-OcCo~\cite{wang2021unsupervised} &89.7 $\pm$ 1.9 &  92.4 $\pm$ 1.6 &  83.9 $\pm$ 1.8 & 89.7 $\pm$ 1.5\\
    PointNet-CrossPoint~\cite{afham2022crosspoint} &90.9 $\pm$ 4.8 &  93.5 $\pm$ 4.4 &  84.6 $\pm$ 4.7 & 90.2 $\pm$ 2.2\\
    DGCNN~\cite{wang2019dynamic} &31.6 $\pm$ 2.8 &  40.8 $\pm$ 4.6&  19.9 $\pm$  2.1& 16.9 $\pm$ 1.5\\
    DGCNN-CrossPoint~\cite{afham2022crosspoint} &92.5 $\pm$ 3.0 & 94.9 $\pm$ 2.1 &83.6 $\pm$ 5.3 &87.9 $\pm$ 4.2\\
    \midrule[0.6pt]
    \multicolumn{5}{c}{\textit{with Self-Supervised Representation Learning} ({\scshape Full})}\\
    \midrule[0.6pt]
    Transformer~\cite{vaswani2017attention} & 87.8 $\pm$ 5.2& 93.3 $\pm$ 4.3 & 84.6 $\pm$ 5.5 & 89.4 $\pm$ 6.3\\
    OcCo~\cite{wang2021unsupervised} & 94.0 $\pm$ 3.6& 95.9 $\pm$ 2.3 & 89.4 $\pm$ 5.1 & 92.4 $\pm$ 4.6\\
    Point-BERT~\cite{PointBERT} & 94.6 $\pm$ 3.1 & 96.3 $\pm$ 2.7 & 91.0 $\pm$ 5.4 & 92.7 $\pm$ 5.1\\
    MaskPoint~\cite{MaskPoint} & 95.0 $\pm$ 3.7 & 97.2 $\pm$ 1.7 & 91.4 $\pm$ \textbf{4.0} & 93.4 $\pm$ 3.5\\
    Point-MAE~\cite{PointMAE} & \underline{96.3} $\pm$ 2.5&\underline{97.8} $\pm$ 1.8 & \underline{92.6} $\pm$ 4.1 & \underline{95.0} $\pm$ \textbf{3.0}\\
    \rowcolor{linecolor}Point-CMAE (Ours) & \textbf{96.7} $\pm$ \textbf{2.2} & \textbf{98.0} $\pm$ \textbf{0.9} & \textbf{92.7} $\pm$ 4.4 & \textbf{95.3} $\pm$ 3.3 \\
    \midrule[0.6pt]
    \multicolumn{5}{c}{\textit{with Hierarchical ViTs/Multi-Modal/Post-Process Self-Supervised Learning }({\scshape Full})}\\
    \midrule[0.6pt]
    Point-M2AE~\cite{zhang2022point} & 96.8 $\pm$ 1.8 & 98.3 $\pm$ 1.4 & 92.3 $\pm$ 4.5 & 95.0 $\pm$ 3.0 \\
    Joint-MAE~\cite{guo2023joint} & 96.7 $\pm$ 2.2 & 97.9 $\pm$ 1.8 & 92.6 $\pm$ 3.7 & 95.1 $\pm$ 2.6 \\
    Point-GPT~\cite{chen2024pointgpt} & 96.8 $\pm$ 2.0 & 98.6 $\pm$ 1.1 & 92.6 $\pm$ 4.6 & 95.2 $\pm$ 3.4 \\
    ACT~\cite{dong2022autoencoders} & 96.8 $\pm$ 2.3 & 98.0 $\pm$ 1.4 & 93.3 $\pm$ 4.0 & 95.6 $\pm$ 2.8 \\
    \midrule[0.1pt]
    \midrule[0.1pt]
    \multicolumn{5}{c}{\textit{with Self-Supervised Representation Learning} ({\scshape Mlp-Linear})}\\
    \midrule[0.6pt]
    Point-MAE & \textbf{91.1} $\pm$ 5.6 & 91.7 $\pm$ 4.0 & 83.5 $\pm$ 6.1 & 89.7 $\pm$ \textbf{4.1}\\
    \rowcolor{linecolor}Point-CMAE (Ours) & 90.4 $\pm$ \textbf{4.2} & {\bf 94.1 }$\pm$ \textbf{3.9} & {\bf 89.2 }$\pm$ \textbf{5.5} & {\bf 92.3 }$\pm$ 4.5\\
    \midrule[0.6pt]
    \multicolumn{5}{c}{\textit{with Self-Supervised Representation Learning} ({\scshape Mlp-$3$})}\\
    \midrule[0.6pt]
    Point-MAE & 95.0 $\pm$ \textbf{2.8} & 96.7 $\pm$ 2.4 & 90.6 $\pm$ 4.7 & 93.8 $\pm$ 5.0\\
    \rowcolor{linecolor}Point-CMAE (Ours) & \textbf{95.9} $\pm$ 3.1 & \textbf{97.5} $\pm$  \textbf{2.0} & {\bf91.3} $\pm$ {\bf4.6}& {\bf 94.4} $\pm$ \textbf{3.7}\\
    \bottomrule[0.95pt]
    \end{tabular}
    }
\vspace{-5mm}
\end{table}

%% file: tables/partseg_shapenetpart.tex
\begin{table}[t!]
    \centering
    \caption{Part segmentation on \textbf{ShapeNetPart}. The class mIoU (mIoU$_C$) and the instance mIoU (mIoU$_I$) are reported, with model parameters \#P and FLOPs \#F.}
    \label{tab:partseg}
    \setlength\tabcolsep{7pt}
    \setlength{\extrarowheight}{0.3pt}
    \scalebox{0.87}{
    \begin{tabular}{lccc}
    \toprule[0.95pt]
    Method & mIoU$_C$ (\%) $\uparrow$ & mIoU$_I$ (\%) $\uparrow$ & \#P (M) $\downarrow$ \\
    \midrule[0.6pt]
    \multicolumn{4}{c}{\textit{Supervised Representation Learning}}\\
    \midrule[0.6pt]
    PointNet~\cite{PointNet}  & 80.4 & 83.7 & 3.6  \\
    PointNet++~\cite{PointNet++} & 81.9 & 85.1 & {1.0} \\
    DGCNN~\cite{DGCNN} & 82.3 & 85.2 & 1.3 \\
    Transformer~\cite{vaswani2017attention} & 83.4 & 85.1 & 22.1  \\
    PointMLP~\cite{ma2021rethinking} & 84.6 & 86.1 & -\\
    \midrule[0.6pt]
    \multicolumn{4}{c}{\textit{with Self-Supervised Representation Learning} }\\
    \midrule[0.6pt]
    Transformer~\cite{vaswani2017attention} & 83.4 & 85.1 & 22.1 \\
    OcCo~\cite{wang2021unsupervised} & 83.4 & 84.7 & 22.1 \\
    PointContrast~\cite{xie2020pointcontrast} & - & 85.1 & 37.9 \\
    CrossPoint~\cite{afham2022crosspoint} & - & 85.5 & - \\
    Point-BERT~\cite{PointBERT} & 84.1 & 85.6 & 22.1  \\
    MaskPoint~\cite{MaskPoint} & 84.6 & \underline{86.0} & 22.1 \\
    Point-MAE~\cite{PointMAE} & {84.2} & \textbf{86.1} & 22.1  \\
    \rowcolor{linecolor}\textbf{Point-CMAE (Ours)} & \textbf{84.9} & \underline{86.0} & {22.1} \\
    \midrule[0.6pt]
    \multicolumn{4}{c}{\textit{with Hierarchical ViTs / Multi-Modal/Post-Process Self-Supervised Learning} }\\
    \midrule[0.6pt]
    Point-M2AE~\cite{zhang2022point} & 84.8 & 86.5 & 12.8 \\
    PointGPT-L~\cite{chen2024pointgpt} & 84.8 & 86.6 & 29.2\\
    ACT~\cite{dong2022autoencoders} & 84.7 & 86.1 & 22.1 \\ 
    Recon~\cite{qi2023contrast} & 84.8 & 86.4 & 43.6 \\
    \bottomrule[0.95pt]
    \end{tabular}
    }
\vspace{-5mm}
\end{table}

%% file: tables/ablation.tex
\begin{table}[t]
\centering
\caption{The effect of each component on ScanObjectNN~\cite{uy2019revisiting} and ShapeNetPart~\cite{yi2016scalable} datasets for classification (OA) and part segmentation (mIoU$_{C}$).}
\label{tab:ablation}
\setlength\tabcolsep{5pt}
\setlength{\extrarowheight}{1pt}
\scalebox{0.8}{
\begin{tabular}{lcccccc}
\toprule[0.95pt]
    Methods & Dual-masking & Same Encoder & Same decoder & Contrastive & OA(\%) $\uparrow$  & mIoU$_C$ (\%) $\uparrow$ \\
    \midrule
    Baseline~\cite{PointMAE} & $\times$ & - & - & - & 85.18 & 84.20 \\
    \midrule[0.6pt]
    (a) & \checkmark & $\times$ & $\times$ & $\times$ & 85.51 & 84.32 \\
    (b) & \checkmark & $\times$ & $\times$ & \checkmark & 85.60 & 84.45 \\
    (c) & \checkmark & $\times$ & \checkmark & $\times$ & 85.23 & 84.33 \\
    (d) & \checkmark & $\times$ & \checkmark & \checkmark & 85.32 & 84.51 \\
    \midrule[0.6pt]
    (e) & \checkmark & \checkmark & $\times$ & $\times$ & 85.70 & 84.65 \\
    \rowcolor{linecolor}(f) & \checkmark & \checkmark & $\times$ & \checkmark & 85.95 & 84.85 \\
    (g) & \checkmark & \checkmark & \checkmark & $\times$ & 85.43 & 84.50 \\
    (h) & \checkmark & \checkmark & \checkmark & \checkmark & 85.37 & 84.43 \\
\bottomrule[0.95pt]
\end{tabular}
}
\vspace{-5mm}
\end{table}

%% file: sections/5_conclusion.tex
\section{Conclusion}
\label{sec:conclusion}
We propose Point-CMAE, a self-supervised method that integrates the MAE pre-training paradigm with explicit contrastive properties for point clouds. Specifically, we experimentally demonstrate that directly combining classic contrastive learning with generative MAE degrades the SSL performance. To address this, we propose a simple dual masking strategy that effectively introduces explicit contrastive properties. Then the feature level contrastive constraint is applied which ensures our Point-CMAE achieves significant improvements compared to the baseline method and even outperforms other state-of-the-art methods that incorporate complex hierarchical architectures, post-training techniques, or multi-modal information. Additionally, our findings indicate that masking plays a significant role in MAE-based SSL, highlighting the importance of careful attention to masking strategies, especially for point clouds.

%% file: sections/appendix.tex
\section{Additional Implementation Details}

\noindent\textbf{Masked Point Modeling Reconstruction Loss:}~  
For the masked autoencoder (MAE) loss (\ie, $ \mathcal{L}_{\text{recon}}$), we use the $\ell_2$ Chamfer-Distance~\cite{fan2017point} following~\cite{PointMAE}.
Let $\mathcal{R} \equiv \operatorname{RP}(g_{\phi}^{1}(z_1))$ and $\mathcal{G} \equiv X$ be the reconstructed point clouds and ground truth point clouds, respectively. The reconstruction loss $ \mathcal{L}_{\text{recon}}$ can be written as:
\begin{equation}\label{eq:mpm}
    \mathcal{L}_{\text{recon}} =
    \sum_{} 
    \left[
        \frac{1}{|\mathcal{R}|} \sum_{{re}\in\mathcal{R}}\mathop{\min}\limits_{gt\in\mathcal{G}} \|re-gt\|^2_2 +
        \sum_{gt\in\mathcal{G}}\mathop{\min}\limits_{re\in\mathcal{R}} \|re-gt\|^2_2
    \right].
\end{equation}
\input{tables/hyperparams}

\noindent\textbf{Detailed Training configurations:}~ We also provide the detailed training recipes for both the pre-training and downstream fine-tuning of our Point-CMAE in Tab.~\ref{tab:hyper_params}. Similarly to ACT~\cite{dong2022autoencoders} or Recon~\cite{qi2023contrast}, we adopt two kinds of augmentations (\ie, Scale\&Translate and Rotation) in this work for pre-training and classification on ShapeNet~\cite{chang2015shapenet} and ScanObjectNN~\cite{uy2019revisiting} datasets, respectively.  

\section{Limitation and Future Works}
While the proposed Point-CMAE achieves competitive results across various downstream tasks, it does not incorporate the multi-modal information commonly used in recent research. Future work should explore extending Point-CMAE to integrate additional modalities, such as images and depth maps, to potentially enhance its performance. Additionally, the current work employs the same mask ratio for both masks. Investigating the impact of varying the mask ratios could provide deeper insights into the invariance properties of Vision Transformers (ViTs) for point cloud representation learning. This could further refine our understanding and improve the model's robustness and generalization capabilities.

\section{Broader Impact}
Our research introduces Point-CMAE, an innovative approach integrating contrastive learning with masked autoencoder pre-training for Vision Transformers in 3D point cloud data. This advancement enhances 3D object recognition and segmentation, improving applications like autonomous driving and robotics while promoting data efficiency, and making robust models feasible even with limited labeled data. The techniques developed can inspire innovations in other domains, fostering cross-disciplinary advancements. By releasing our code, we contribute to the open-source community, facilitating collaboration and reproducibility. Point-CMAE also serves as a valuable educational resource, while its ethical application can enhance societal benefits.

%% file: tables/hyperparams.tex
\begin{center}
\begin{table}[t]
\caption{\textbf{Training recipes for pre-training and downstream fine-tuning}.}
\vspace{-5mm}
\label{tab:hyper_params}
\vskip 0.10in
\centering
\scalebox{0.92}{
\begin{tabular}{lcccc}
 & \texttt{Pre-training} & \multicolumn{2}{c}{\texttt{Classification}} & \texttt{Segmentation}\\
 \toprule[0.95pt]
 Config & ShapeNet~\cite{chang2015shapenet} & ScanObjectNN~\cite{uy2019revisiting} & ModelNet~\cite{wu20153d} & ShapeNetPart~\cite{yi2016scalable}\\
 \midrule[0.6pt]
 optimizer & AdamW & AdamW & AdamW & AdamW\\
 learning rate & 1e-3 & 5e-4 & 5e-4 & 1e-4 \\
 weight decay & 5e-2 & 5e-2 & 5e-2 & 5e-2 \\
 learning rate scheduler & cosine & cosine & cosine & cosine \\
 training epochs & 300 & 300 & 300 & 300\\
 warmup epochs & 10 & 10 & 10 & 10\\
 batch size & 128 & 32 & 32 & 16\\
 drop path rate & 0.1 & 0.1 & 0.1 & 0.1 \\
 \midrule[0.6pt]
 number of points & 1024 & 2048 & 1024 & 2048 \\
 number of point patches & 64 & 128 & 64 & 128 \\
 point patch size & 32 & 32 & 32 & 32 \\
 \midrule[0.6pt]
 augmentation1 & Scale\&Trans & Scale\&Trans & Scale\&Trans & - \\
  augmentation2 & Rotation & Rotation & Scale\&Trans & - \\
 \midrule[0.6pt]
 GPU device & 1 A100 (40G) & 1 A100 (40G) & 1 A100 (40G) & 1 A100 (40G) \\
\bottomrule[0.95pt]
\end{tabular}
}
\end{table}
\end{center}